\documentclass[letter,onecolumn,11pt]{article}
\usepackage{fullpage}
\usepackage{times}
\usepackage{url}
\usepackage{latexsym}
\usepackage{amsmath,amsfonts,amssymb}
\usepackage{tikz}
\usepackage{color,umoline,correct}
\usepackage{authblk,url}

\usepackage{subfigure}
\graphicspath{{./}}
\begin{document}

\title{Linguistic Geometries for Unsupervised Dimensionality Reduction}
\author{Yi Mao\textsuperscript{*}}
\author{Krishnakumar Balasubramanian}
\author{Guy Lebanon}
\affil{
	School of Computational Science \& Engineering \\ 
	College of Computing \\
	Georgia Institute of Technology \\ Atlanta, Georgia
}
\date{\today}
\maketitle
\thispagestyle{empty}
\let\oldthefootnote\thefootnote
\renewcommand{\thefootnote}{\fnsymbol{footnote}}
\footnotetext[1]{To whom correspondence should be addressed. Email: \url{yi.mao@cc.gatech.edu}}
\let\thefootnote\oldthefootnote

\maketitle
\begin{abstract}
Text documents are complex high dimensional objects. To effectively visualize such data it is important to reduce its dimensionality and visualize the low dimensional embedding as a 2-D or 3-D scatter   plot. In this paper we explore dimensionality reduction methods that draw upon domain knowledge in order to achieve a better low dimensional embedding and visualization of documents. We consider the use of geometries specified manually by an expert, geometries derived automatically from corpus statistics, and geometries computed from linguistic resources.
\end{abstract}

\section{Introduction} \label{sec:intro}

Visual document analysis systems such as IN-SPIRE have demonstrated their applicability in managing large text corpora, identifying topics within a document and quickly identifying a set of relevant documents by visual exploration. The success of such systems depends on several factors with the most important one being the quality of the dimensionality reduction. This is obvious as visual exploration can be made possible only when the dimensionality reduction preserves the structure of the original space, i.e., documents that convey similar topics are mapped to nearby regions in the low dimensional 2D or 3D space.

Standard dimensionality reduction methods such as principal component analysis (PCA), locally linear embedding (LLE) \cite{Roweis2000}, or t-distributed stochastic neighbor embedding (t-SNE) \cite{van2008visualizing} take as input a set of feature vectors such as bag of words or tf vectors. An obvious drawback of such an approach is that such methods ignore the textual nature of documents and instead consider the vocabulary words $V=\{v_1,\ldots,v_n\}$ as abstract orthogonal dimensions that are unrelated to each other. In this paper we introduce a general technique for incorporating domain knowledge into dimensionality reduction for text documents. In contrast to several recent alternatives, our technique is completely  unsupervised and does not require any labeled data.

We focus on the following type of non-Euclidean geometry where the distance between document $x$ and $y$ is defined as
\begin{align} \label{eq:dist}
d_T(x,y) = \sqrt{(x-y)^{\top} T (x-y)}.
\end{align}
Here $T \in \mathbb{R}^{n\times n}$ is a symmetric positive semidefinite matrix, and we assume that documents $x,y$ are represented as term-frequency (tf) column vectors. Since $T$ can always be written as $H^{\top}H$ for some matrix $H \in \mathbb{R}^{m \times n}$ where $m \leq n$, an equivalent but sometimes more intuitive interpretation of \eqref{eq:dist} is to compose the mapping $x\mapsto Hx$ with the Euclidean geometry
\begin{align} \label{eq:distAlt}
d_T(x,y) & = d_I(Hx,Hy) = \|Hx-Hy\|_2.
\end{align}
We can view $T$ as encoding the semantic similarity between pairs of words. When $H$ is a square matrix, it smoothes the tf vector $x$ by mapping observed words to unobserved related words. Alternatively, if $m$, the number of rows of $H$, equals to the number of existing topics, the mapping can be viewed as describing a document as a mixture of such topics. Therefore, the geometry realized by \eqref{eq:dist} or \eqref{eq:distAlt} may be used to derive novel dimensionality reduction methods that are customized to text in general and to specific text domains in particular. The main challenge is to obtain the matrices $H$ or $T$ that describe the relationship among vocabulary words appropriately.

We consider obtaining $H$ or $T$ using three general types of domain knowledge. The first corresponds to manual specification of the semantic relationship among words. The second corresponds to analyzing the relationship between different words using corpus statistics. The third corresponds to knowledge obtained from linguistic resources. In some cases, $T$ might be easier be obtain than $H$. Whether to specify $H$ directly or indirectly through $T$ depends on the knowledge type and is discussed in detail in Section \ref{sec:domain}.

We investigate the performance of the proposed dimensionality reduction methods for three text domains: sentiment visualization for movie reviews, topic visualization for newsgroup discussion articles, and visual exploration of ACL papers. In each of these domains we compare several different domain dependent geometries and show that they outperform popular state-of-the-art techniques. Generally speaking, we observe that geometries obtained from corpus statistics are superior to manually constructed geometries and to geometries derived from standard linguistic resources such as Word-Net. We also demonstrate effective ways to combine different types of domain knowledge and show how such combinations significantly outperform any of the domain knowledge types in isolation. All the techniques mentioned in this paper are unsupervised, making use of labels only for evaluation purposes.

\section{Related Work} \label{sec:related}

Despite having a long history, dimensionality reduction is still an active research area. Broadly speaking, dimensionality reduction methods may be classified to projective or manifold based \cite{chrismsrTR}. The first projects data onto a linear subspace (e.g., PCA and canonical correlation analysis) while the second traces a low dimensional nonlinear manifold on which data lies (e.g., multidimensional scaling, isomap, Laplacian eigenmaps, LLE and t-SNE). The use of dimensionality reduction for text documents is surveyed by \cite{nvacbook} who also describe current homeland security applications.

Dimensionality reduction is closely related to metric learning. \cite{Xing2003b} is one of the earliest papers that focus on learning metrics of the form \eqref{eq:dist}. In particular they try to learn matrix $T$ in an supervised way by expressing relationships between pairs of samples. Representative paper on unsupervised metric learning for text documents is \cite{Lebanon2006a} which learns a metric on the simplex based on the geometric volume of the data.

We focus in this paper on visualizing a corpus of text documents using a 2-D scatter plot. While this is perhaps the most popular and practical text visualization technique, other methods such as \cite{Spoerri93}, \cite{Hearst1997}, \cite{Havre2002}, \cite{Paley2002}, \cite{Blei2003}, \cite{Mao2007} exist. It is conceivable that the techniques developed in this paper may be ported to enhance these alternative visualization methods as well. 
\section{Non-Euclidean Geometries } \label{sec:geometries}

Dimensionality reduction methods often assume, either explicitly
or implicitly, Euclidean geometry. For example, PCA minimizes the
reconstruction error for a family of Euclidean projections. LLE
uses the Euclidean geometry as a local metric. t-SNE is based on a
neighborhood structure, determined again by the Euclidean
geometry. The generic nature of the Euclidean geometry makes it
somewhat unsuitable for visualizing text documents as the
relationship between words conflicts with Euclidean orthogonality.
We consider in this paper several alternative geometries of the
form \eqref{eq:dist} or \eqref{eq:distAlt} which are more suited
for text and compare their effectiveness in visualizing
documents.

As mentioned in Section~\ref{sec:intro} $H$ smoothes the tf vector $x$ by mapping the observed words into observed and non-observed (but related) words. Decomposing $H = R \times D$ into a product of a Markov
morphism\footnote{a non-negative matrix whose columns sum to 1
\cite{Chentsov1982}} $R\in\mathbb{R}^{n\times n}$ and a
non-negative diagonal matrix $D\in\mathbb{R}^{n\times n}$, we see
that the matrix $H$ plays two roles: blending related vocabulary
words (realized by $R$) and emphasizing some words over others
(realized by $D$). The $j$-th column of $R$ stochastically
smoothes word $w_j$ into related words $w_i$ where the amount of
smoothing is determined by $R_{ij}$. Intuitively $R_{ij}$ is high
if $w_i,w_j$ are similar and $0$ if they are unrelated.
The role of the matrix $D$ is to emphasize some words over others.
For example, $D_{ii}$ values corresponding to content words may
be higher than values corresponding to stop words or less
important words.

\begin{figure}\centering
  \begin{align*}
    \begin{pmatrix}
      0.8 & 0.1 & 0.1 & 0 & 0\\
      0.1 & 0.8 & 0.1 & 0 & 0\\
      0.1 & 0.1 & 0.8 & 0 & 0\\
      0   & 0   & 0   & 0.9 & 0.1\\
      0   & 0   & 0   & 0.1 & 0.9\\
    \end{pmatrix}
    \begin{pmatrix}
      5 & 0 & 0 & 0 & 0\\
      0 & 5 & 0 & 0 & 0\\
      0 & 0 & 5 & 0 & 0\\
      0   & 0   & 0   & 3 & 0\\
      0   & 0   & 0   & 0 & 3\\
    \end{pmatrix}
  \end{align*}
  \caption{An example of a decomposition $H=RD$ in the case of two word clusters $\{v_1,v_2,v_3\}$, $\{v_4,v_5\}$. The block diagonal elements in $R$ represent the fact that words are mostly mapped to themselves, but sometimes are mapped to other words in the same cluster. The diagonal matrix represents the fact that the first cluster is somewhat more important than the second cluster for the purposes of dimensionality
  reduction.}
  \label{fig:matrix}
\end{figure}

It is instructive to examine the matrices $R$ and $D$ in the case
where the vocabulary words cluster according to some meaningful
way. Figure \ref{fig:matrix} gives an example where vocabulary
words form two clusters. The matrix $R$ may become block-diagonal
with non-zero elements occupying diagonal blocks representing
within-cluster word blending, i.e., words within each cluster are
interchangeable to some degree. The diagonal matrix $D$ represents
the importance of different clusters. The word clusters are formed
with respect to the visualization task at hand. For example, in
the case of visualizing the sentiment content of reviews we may
have word clusters labeled as ``positive sentiment words'',
``negative sentiment words'' and ``objective words''. 
In general, the matrices $R,D$ may be defined based on the
language or may be specific to document domain and visualization
purpose. It is reasonable to expect that the words emphasized for
visualizing topics in news stories might be different than the
words emphasized for visualizing writing styles or sentiment
content.

The above discussion remains valid when $H \in \mathbb{R}^{m
\times n}$ for $m$ being the number of topics in the set of
documents. In fact, the $j$-th column of $R$ now stochastically
maps word $j$ to related topics $i$.

Applying the geometry \eqref{eq:dist} or \eqref{eq:distAlt} to
dimensionality reduction is easily accomplished by first mapping
documents $x\mapsto Hx$ and proceeding with standard
dimensionality reduction techniques such as PCA or t-SNE. The
resulting dimensionality reduction is Euclidean in the transformed
space but non-Euclidean in the original space.

In many cases, the vocabulary contains tens of thousands of words
or more making the specification of the matrices $R,D$ a
complicated and error prone task. We describe in the next section
several techniques for specifying $R,D$ in practice. Note, even if
in some cases $R,D$ are obtained indirectly by decomposing $T$
into $H^\top H$, the discussion of the role of $R,D$ is still of
importance as the matrices can be used to come up word clusters
whose quality may be evaluated manually based on the visualization
task at hand.

\section{Domain Knowledge} \label{sec:domain}

We consider four different techniques for obtaining the
transformation matrix $H$. Each technique approaches in one of two
ways: (1) separately obtain the column stochastic matrix $R$ which
blends different words and the diagonal matrix $D$ which
determines the importance of each word; (2) estimate the semantic
similarity matrix $T$ and decompose it as $H^\top H$. To ensure
that $H$ is a non-negative matrix for it to be interpretable,
non-negativity matrix factorization techniques such as the one in
\cite{Ding05} may be applied.

\subsection*{Method A: Manual Specification}

In this method, an expert user manually specifies the matrices
$(R,D)$ based on his assessment of the relationship among the
vocabulary words. More specifically, the user first constructs a
hierarchical word clustering that may depend on the current text
domain, and then specifies the matrices $(R,D)$ with respect to
the cluster membership of the vocabulary.

Denoting the clusters by $C_1,\ldots,C_r$ (a partition of
$\{v_1,\ldots,v_n\}$), the user specifies $R$ by setting the
values
\begin{align}
R_{ij} &\propto \begin{cases}
\rho_{a},  &\quad i = j, v_i\in C_a \\
\rho_{ab}, &\quad i \neq j, v_i\in C_a, v_j\in C_b
\end{cases}
\end{align}
appropriately. The values $\rho_{a}$ and $\rho_{aa}$ together
determine the blending of words from the same cluster. The value
$\rho_{ab}, a\neq b$ captures the semantic similarity between two
clusters. That value may be either computed manually for each pair
of clusters or automatically from the clustering hierarchy (for
example $\rho_{ab}$ can be the minimal number of tree edges
traversed to move from $a$ to $b$). The matrix $R$ is then
normalized appropriately to form a column stochastic matrix. The
matrix $D$ is specified by setting the values
\begin{align}
D_{ii} &= d_{a}, \quad v_i \in C_a
\end{align}
where $d_{a}$ may indicate the importance of word cluster $C_a$ to
the current visualization task. We emphasize that as with the rest
of the methods in this paper, the manual specification is done
without access to labeled data.

Since manual clustering assumes some form of human intervention,
it is reasonable to also consider cases where the user specifies
$(R,D)$ in an interactive manner. That is, the expert specifies an
initial clustering of words and $(R,D)$, views the resulting
visualization and adjusts his selection interactively until he is
satisfied.

\subsection*{Method B: Contextual Diffusion}
An alternative technique which performs substantially better is to consider a transformation based on the similarity between the contextual distributions of the vocabulary words. The contextual distribution of word $v$ is defined as
\begin{align}
  q_v(w)=p(w \text{ appears in } x| v\text{ appears in } x)
\end{align}
where $x$ is a randomly drawn document. In other words $q_v$ is the distribution governing the words appearing in the context of word $v$.

A natural similarity measure between distributions is the Fisher diffusion kernel proposed by \cite{Lafferty2005a}. Applied to contextual distributions as in \cite{Dillon2007} we arrive at the following similarity matrix (where $c>0$)
\begin{align*} T(u,v) = \exp\left(-c\arccos^2\left(\sum_w \sqrt{q_u(w) q_v(w)}\right)\right).
\end{align*}
Intuitively, the word $u$ will be translated or diffused into $v$ depending on the geometric diffusion between the distributions of likely contexts.

We use the following formula to estimate the contextual distribution from a corpus of documents
\begin{align}\label{eq:contDistr}
  q_w(u) &= \sum_{x'} p(u,x'|w) = \sum_{x'} p(u|x',w)p(x'|w) \nonumber\\
         &= \sum_{x'}   \text{tf}(u,x')\frac{\text{tf}(w,x')}{\sum_{x''}\text{tf}(w,x'')} \\
         &= \left(\frac{1}{\sum_{x'}\text{tf}(w,x')}\right)\left(\sum_{x'}\text{tf}(u,x')\text{tf}(w,x')\right)   \nonumber
\end{align}
where $\text{tf}(w,x)$ is the number of times word $w$ appears in document $x$. The contextual distribution $q_w$ or the diffusion matrix $T$ above may be computed in an unsupervised manner without need for labels.

\subsection*{Method C: Web $n$-Grams}
The contextual distribution method above may be computed based on a large collection of text documents such as the Reuters RCV1 dataset. The estimation accuracy of the contextual distribution increases with the number of documents which may not be as large as required. An alternative is to estimate the contextual distributions $q_v$ from the entire $n$-gram content of the web. Taking advantage of the publicly available Google $n$-gram dataset\footnote{The Google $n$-gram dataset contains $n$-gram counts ($n\leq 5$) obtained from Google based on processing over a trillion words of running text.} we can leverage the massive size of the web to construct the similarity matrix $T$. More specifically, we compute the contextual distribution by altering \eqref{eq:contDistr} to account for the proportion of times two words appear together within the $n$-grams (we used $n=3$ in our experiments).

\subsection*{Method D: Word-Net}
The last method we consider uses Word-Net, a standard linguistic resource, to specify the matrix $T$ in \eqref{eq:dist}. This is similar to manual specification (method A) in that it builds on expert knowledge rather than corpus statistics. In contrast to method A, however, Word-Net is a carefully built resource containing more accurate and comprehensive linguistic information such as synonyms, hyponyms and holonyms. On the other hand, its generality puts it at a disadvantage as method A may be used to construct a geometry suited to a specific text domain.

We follow \cite{Budanitsky2001} who compare five similarity measures between words based on Word-Net. In our experiments we use Jiang and Conrath's measure \cite{Jiang1997} (see also \cite{SLP2008})
\begin{align*}
T_{c_1, c_2} = \log \frac{p(c_1)p(c_2)}{2 p(\text{lcs}(c_1,c_2))}
\end{align*}
as it was shown to outperform the others. Above, $\text{lcs}$ stands for the lowest common subsumer, that is, the lowest node in the hierarchy that subsumes (is a hypernym of) both $c_1$ and $c_2$. The quantity $p(c)$ is the probability that a randomly selected word in a corpus is an instance of the synonym set that contains word $c$.

\subsection*{Convex Combinations}
In addition to methods A-D which constitute ``pure methods'' we also consider convex combinations
\begin{align} \label{eq:combinationThm}
  H^* = \sum_i \alpha_i H_i \quad \alpha_i\geq 0, \quad \sum_i \alpha_i=1
\end{align}
where $H_i$ are matrices from methods A-D, and $\alpha$ is a non-negative weight vector which sums to 1. Equation~\ref{eq:combinationThm} allows to combine heterogeneous types of domain knowledge (manually specified such as method A and D and automatically derived such as methods B and C). Doing so leverages their diverse nature and potentially achieving higher performance than each of the methods A-D on its own.

\section{Experiments} \label{sec:experiments}

We evaluated methods A-D and the convex combination method by
experimenting on two datasets from different domains. The first is
the Cornell sentiment scale dataset of movie reviews
\cite{Pang2004}. The visualization in this case focuses on the
sentiment quantity \cite{Pang2002}. For simplicity, we only kept
documents having sentiment level 1 (very bad) and 4 (very good).
Preprocessing included lower-casing, stop words removal, stemming,
and selecting the most frequent 2000 words. Alternative
preprocessing is possible but should not modify the results much
as we focus on comparing alternatives rather than measuring
absolute performance. The second text dataset is 20 newsgroups. It
consists of newsgroup articles from 20 distinct newsgroups and is
meant to demonstrate topic visualization.

To measure the dimensionality reduction quality, we display the data as a scatter plot with different data groups (topics, sentiments) displayed with different markers and colors. Our quantitative evaluation is based on the fact that documents belonging to different groups (topics, sentiments) should be spatially separated in the 2-D space. Specifically, we used the following indices to evaluate different reduction methods and geometries.

\begin{description}
\item[(i)] The weighted intra-inter measure is a standard
clustering quality index that is invariant to non-singular linear
transformations of the embedded data. It equals to $\text{tr}
S_T^{-1} S_W$ where $S_W$ is the within-cluster scatter matrix,
$S_T=S_W+S_B$ is the total scatter matrix, and $S_B$ is the
between-cluster scatter matrix \cite{Duda2001}.

\item[(ii)] The Davies Bouldin index is an alternative to (i) that
is similarly based on the ratio of within-cluster scatter to
between-cluster scatter \cite{dbindex2000}.

\item[(iii)] Classification error rate of a $k$-NN classifier that
applies to data groups in the 2-D embedded space. Despite the fact
that we are not interested in classification per se (otherwise we
would classify in the original high dimensional space), it is an
intuitive and interpretable measure of cluster separation.

\item[(iv)] An alternative to (iii) is to project
the embedded data onto a line which is the direction returned by
applying Fisher's linear discriminant analysis \cite{Duda2001} to
the embedded data. The projected data from each group is fitted to
a Gaussian whose separation is used as a proxy for visualization
quality. In particular, we summarize the separation of the two
Gaussians by measuring the overlap area. While (iii) corresponds
to the performance of a $k$-NN classifier, method (iv) corresponds
to the performance of Fisher's LDA classifier.
\end{description}
Note that the above methods (i)-(iv) make use of labeled information to evaluate visualization quality. The labeled data, however, is not used during the dimensionality reduction stages justifying their unsupervised behavior.

\begin{figure*}
\begin{tikzpicture}
[level distance=20mm,
level 1/.style={sibling distance=22mm},
level 2/.style={sibling distance=9mm},
level 3/.style={sibling distance=8mm},
]
{
\node {\footnotesize Vocabulary}
    child{node{\footnotesize Politics}
    child {node{\tiny Mid East}}
    child {node{\tiny Others}}
    }
    child{node{\footnotesize \begin{tabular}{l}Sci \& \\ Tech\end{tabular}}
    child {node{\tiny Comp}
    child {node{\tiny HW}}
    child {node{\tiny SW}}
    child {node{\tiny GUI}}
    child {node{\tiny Others}}
    }
    child {node{\tiny Med}}
    child {node{\tiny Evol}}
    }
    child{node{\footnotesize Religion}
    child {node{\tiny Christianity}child {node [above =5pt] {\tiny \fbox{\begin {tabular}{l}  bible
 \\ gospel  \\ amen\\ christians \\ santa  \end{tabular}}} edge from parent[draw=none]}}
    child {node{\tiny Others}}
    }
    child{node{\footnotesize History}
    child {node{\tiny People}}
    child {node{\tiny Others}}
    }
    child{node{\footnotesize Sports}
    child {node{\tiny Team Name }}
    child {node{\tiny Others}   child {node [above =5pt] {\tiny \fbox{\begin {tabular}{l}  Canoeing
 \\ catch  \\boxing \\ innings \\ soccer  \end{tabular}}} edge from parent[draw=none]}}
    }
    child{node{\footnotesize Others}}
    child{node{\footnotesize Places}
    child {node{\tiny US} child {node [above =5pt] {\tiny \fbox{\begin {tabular}{l}  Arizona
 \\ francisco  \\ carolina  \\ atlanta \\ austin  \end{tabular}}} edge from parent[draw=none]}  }
    child {node{\tiny EU}}
    child {node{\tiny Asia}
    child {node{\tiny Mid east}}}
    child {node{\tiny Others}}
    };}
\end{tikzpicture}
\caption{Manually specified hierarchical word clustering for the 20 newsgroup domain. The words in the frames are examples of words belonging to several bottom level clusters.} \label{fig:manual}
\end{figure*}
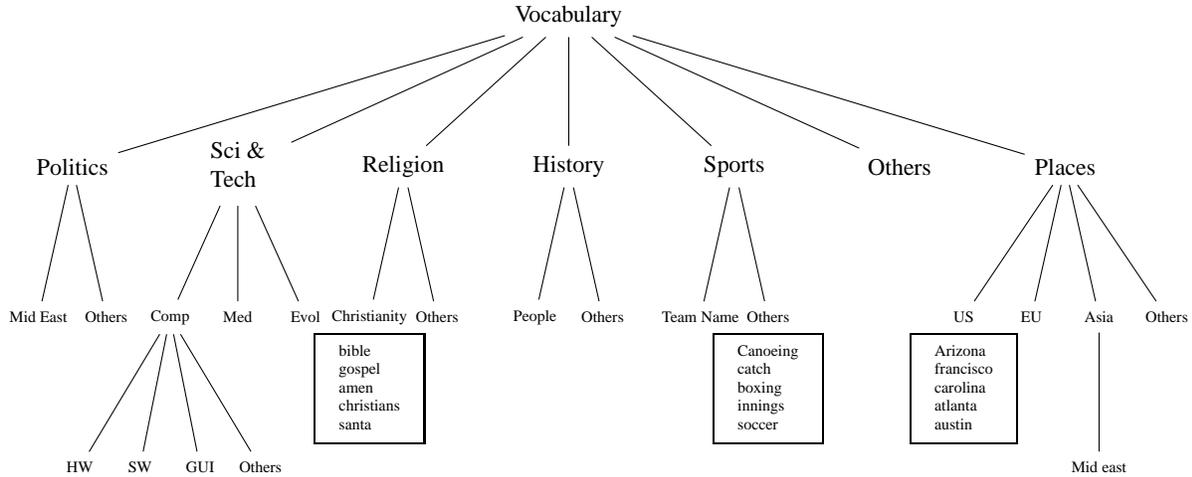

The manual specification of domain knowledge (method A) for the 20
newsgroups domain used matrices $R,D$ that were specified
interactively based on the (manually obtained) word clustering in
Figure~\ref{fig:manual}. In the case of sentiment data the manual
specification consisted of partitioning words into positive,
negative or neutral sentiment based on the General Inquirer
resource\footnote{http://www.wjh.harvard.edu/$\sim$inquirer/}. The
matrix $H$ was completed by assigning large weights ($D_{ii}$) for
negative and positive words and small weights ($D_{ii}$) to
neutral words.

The contextual diffusion
(method B) was computed from a large external corpus (Reuters
RCV1) for the newsgroups domain. For the sentiment domain we used
movie reviews authored by other critics. Google $n$-gram (method
C) provided a truly massive scale resource for estimating the
contextual diffusion. In the case of Word-Net (method D) we used
Ted Pedersen's implementation of Jiang and Conrath's similarity
measure\footnote{http://wn-similarity.sourceforge.net/}. Note, for
method C and D, the resulting matrix $H$ is not domain specific
but rather represents general semantic relationships between
words.

In our experiments below we focused on two dimensionality reduction methods: PCA and t-SNE. PCA is a well known classical method while t-SNE  \cite{van2008visualizing} is a recently proposed technique shown to outperform LLE, CCA, MVU, Isomap, and Laplacian eigenmaps. Indeed it is currently considered state-of-the-art for dimensionality reduction for visualization purposes.

\begin{figure*}
\centering  \subfigure{\fbox{\includegraphics[scale=0.33]{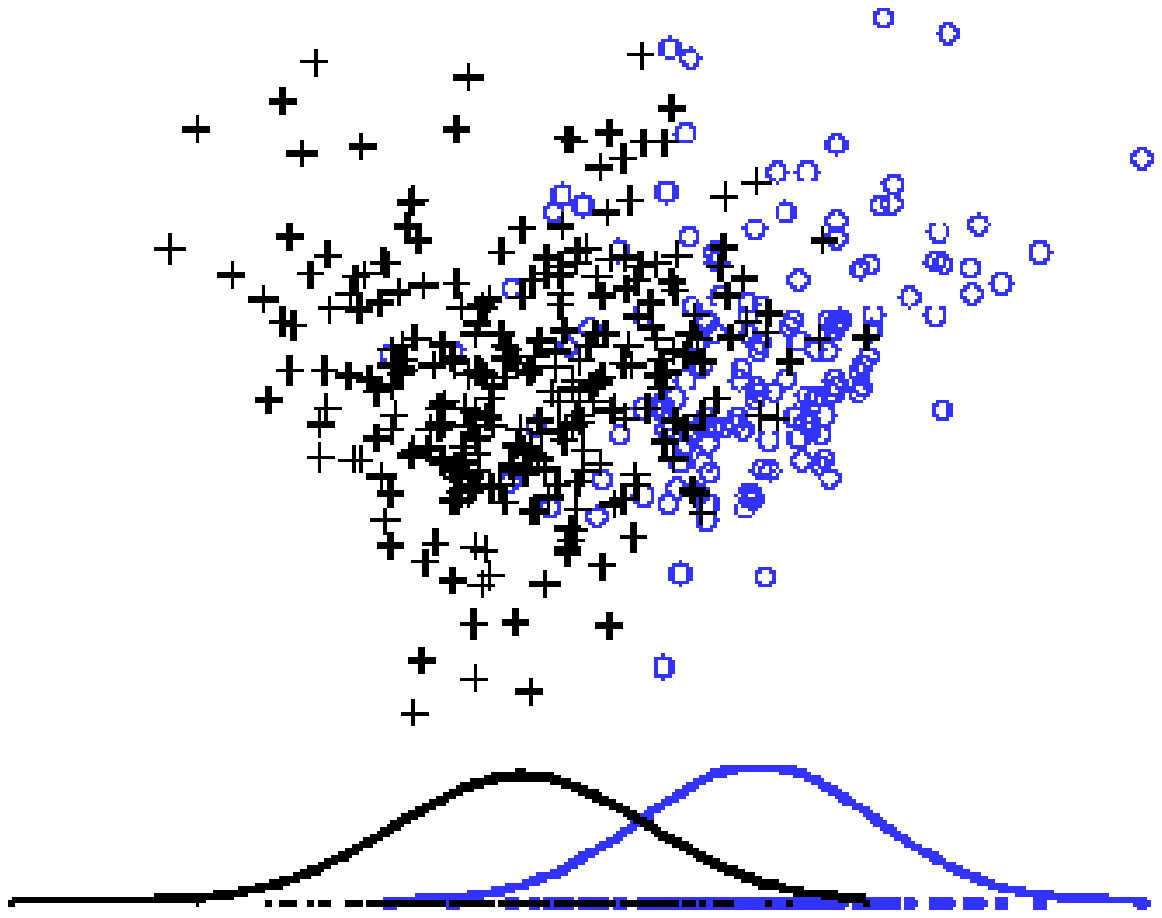}}} \hfil
\subfigure{\fbox{\includegraphics[scale=0.33]{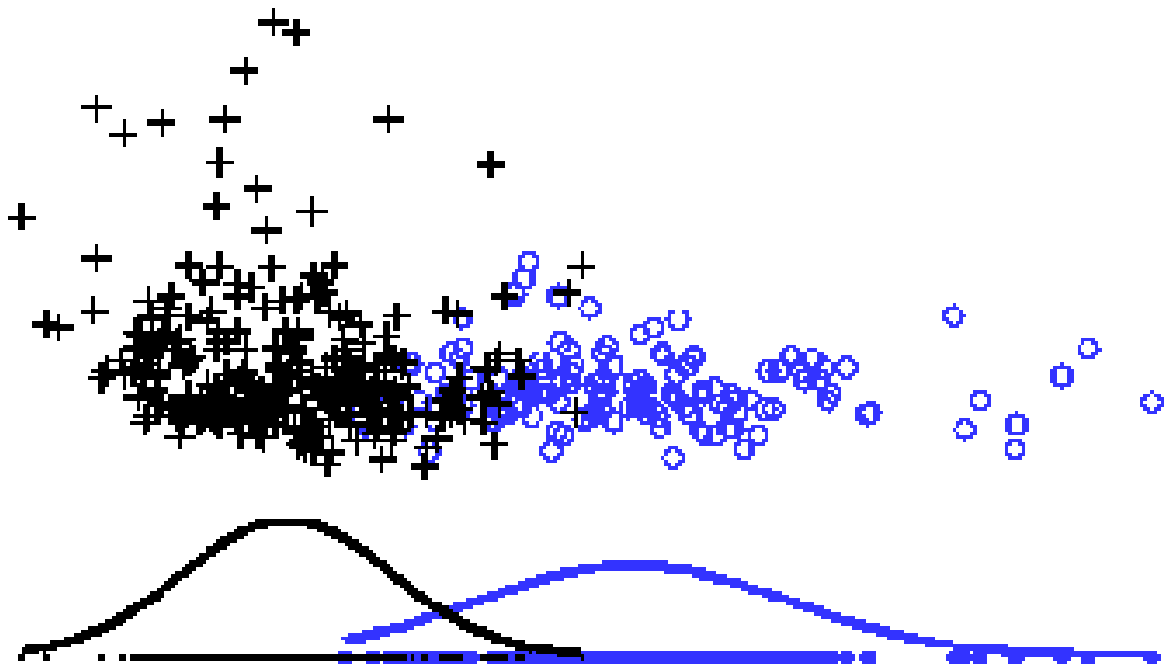}}} \hfil
\subfigure{\fbox{\includegraphics[scale=0.33]{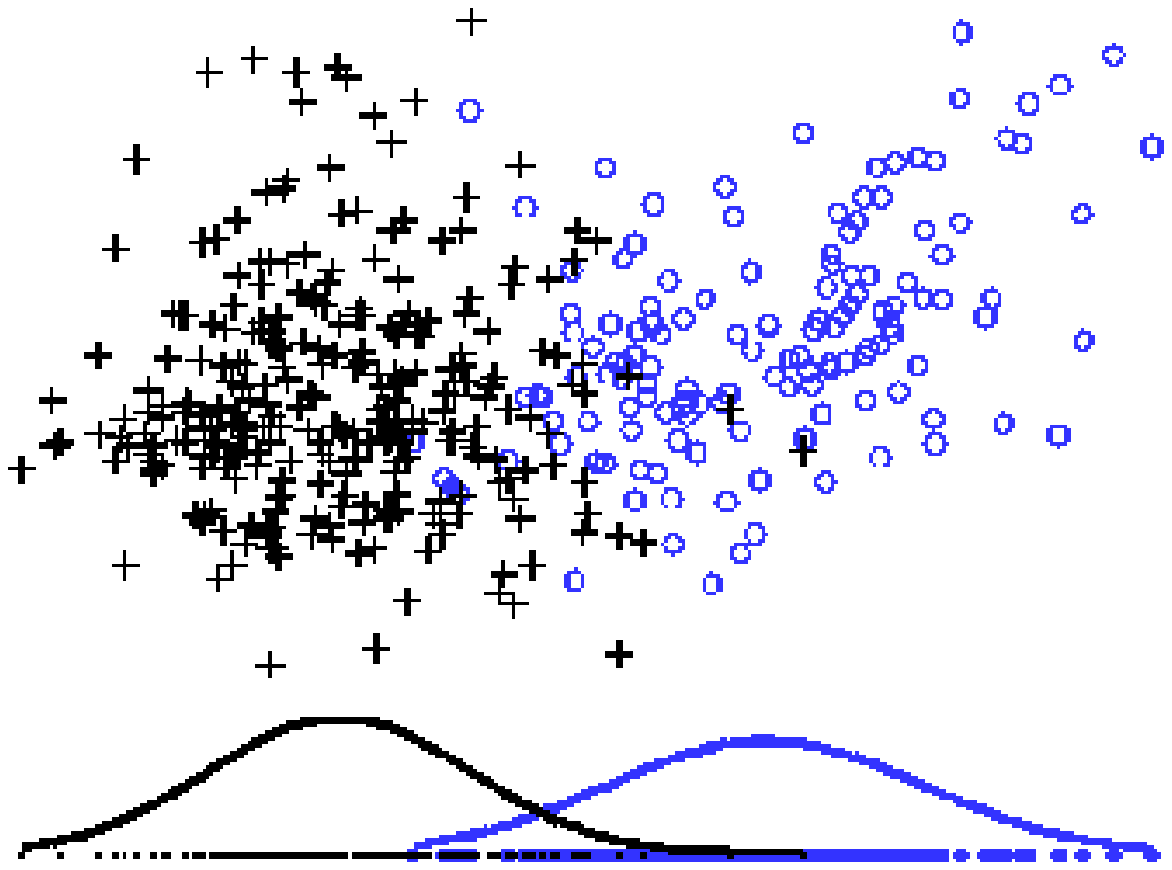}}}\\
\subfigure{\fbox{\includegraphics[scale=0.33]{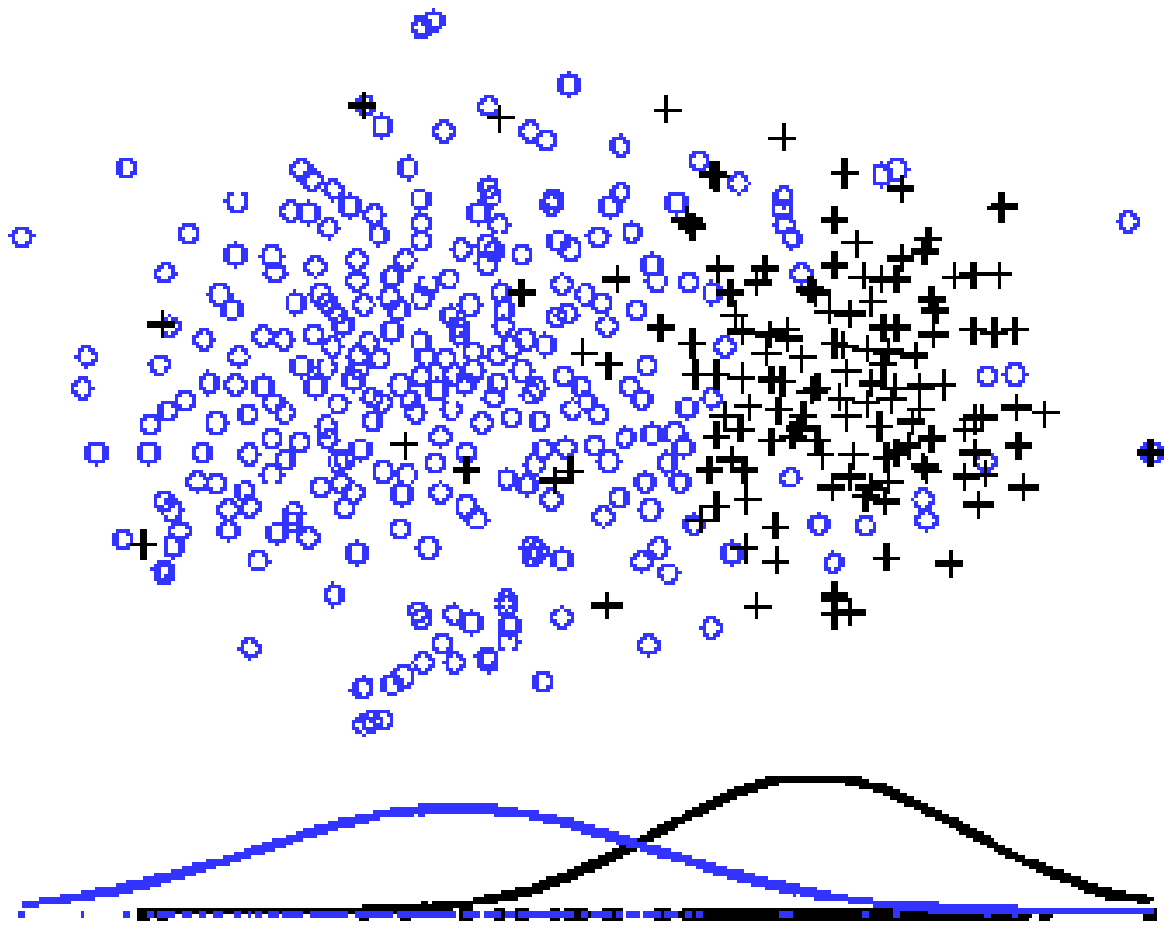}}} \hfil
\subfigure{\fbox{\includegraphics[scale=0.33]{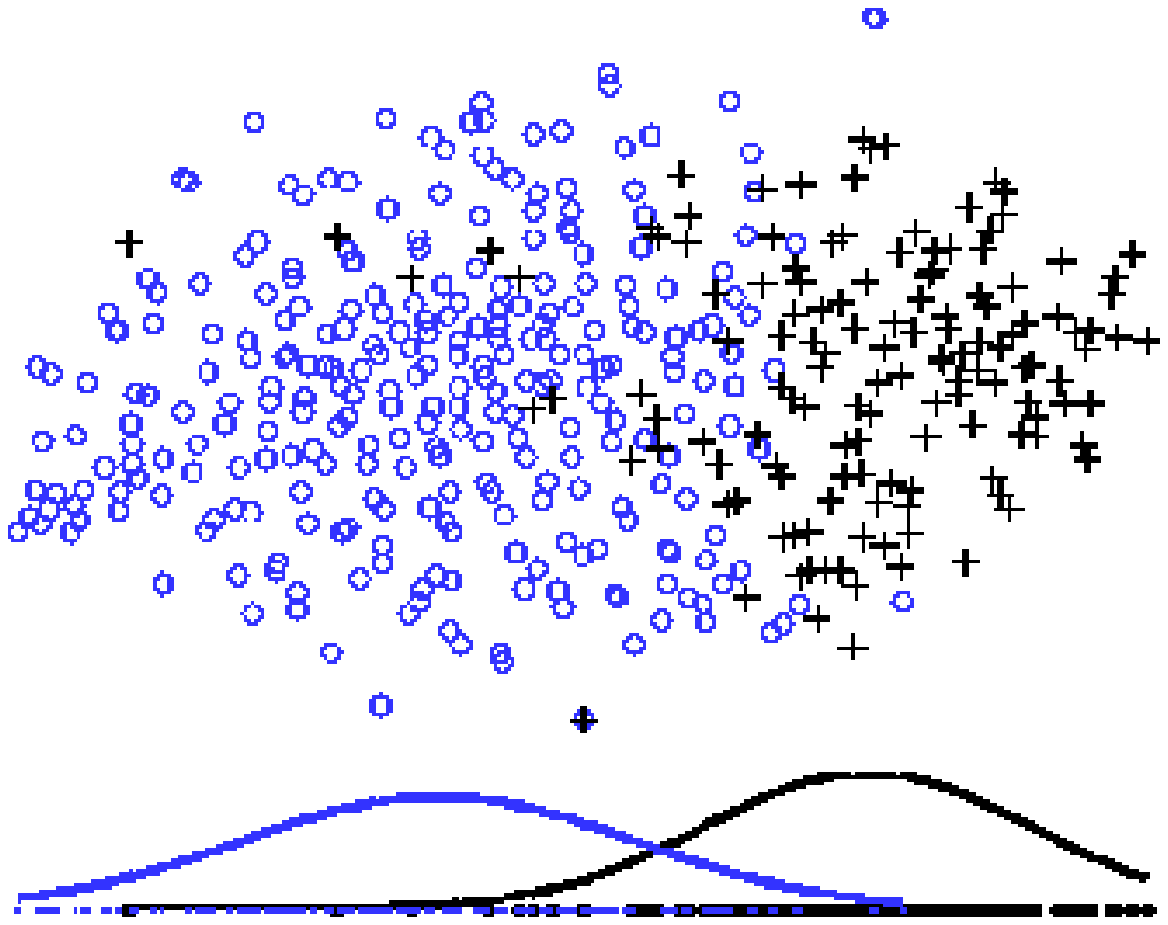}}} \hfil
\subfigure{\fbox{\includegraphics[scale=0.33]{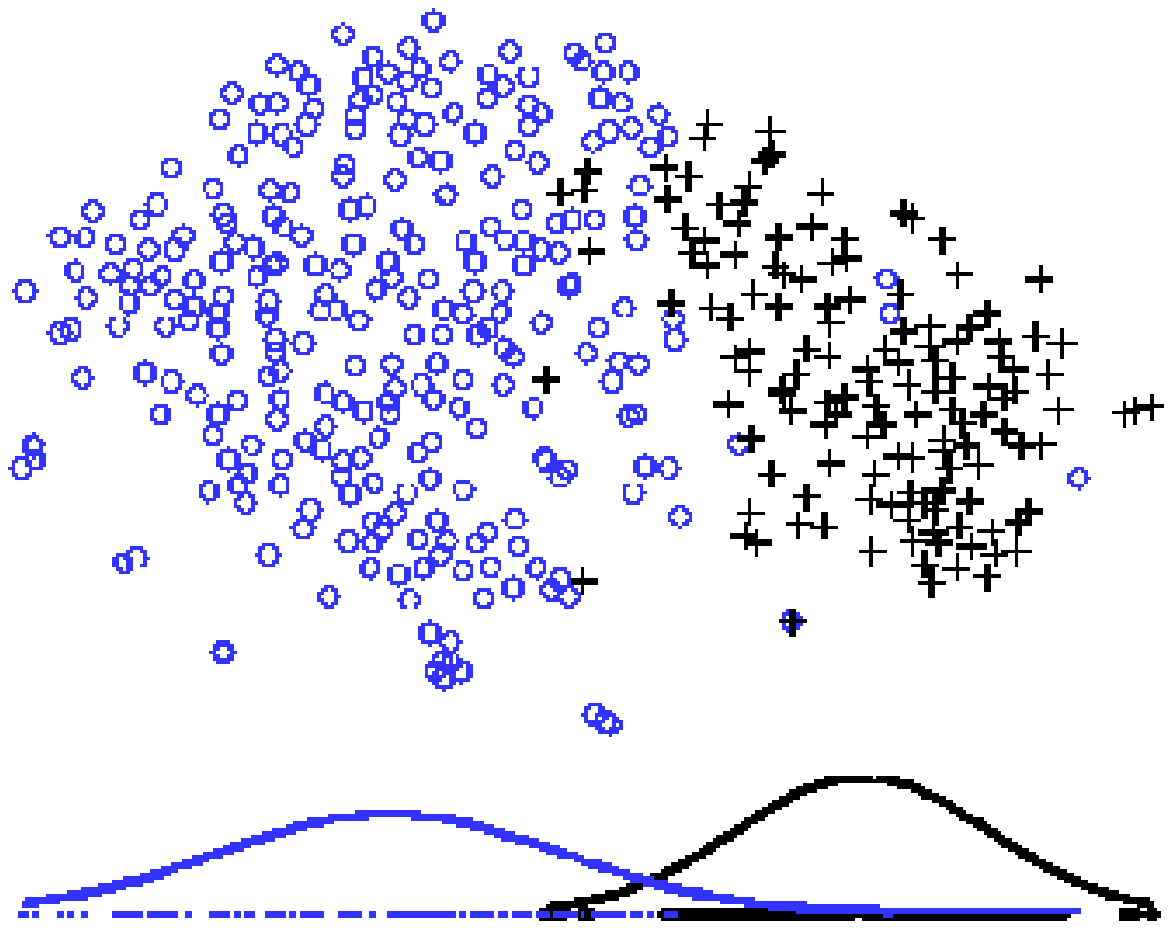}}}\\
\subfigure{\fbox{\includegraphics[scale=0.33]{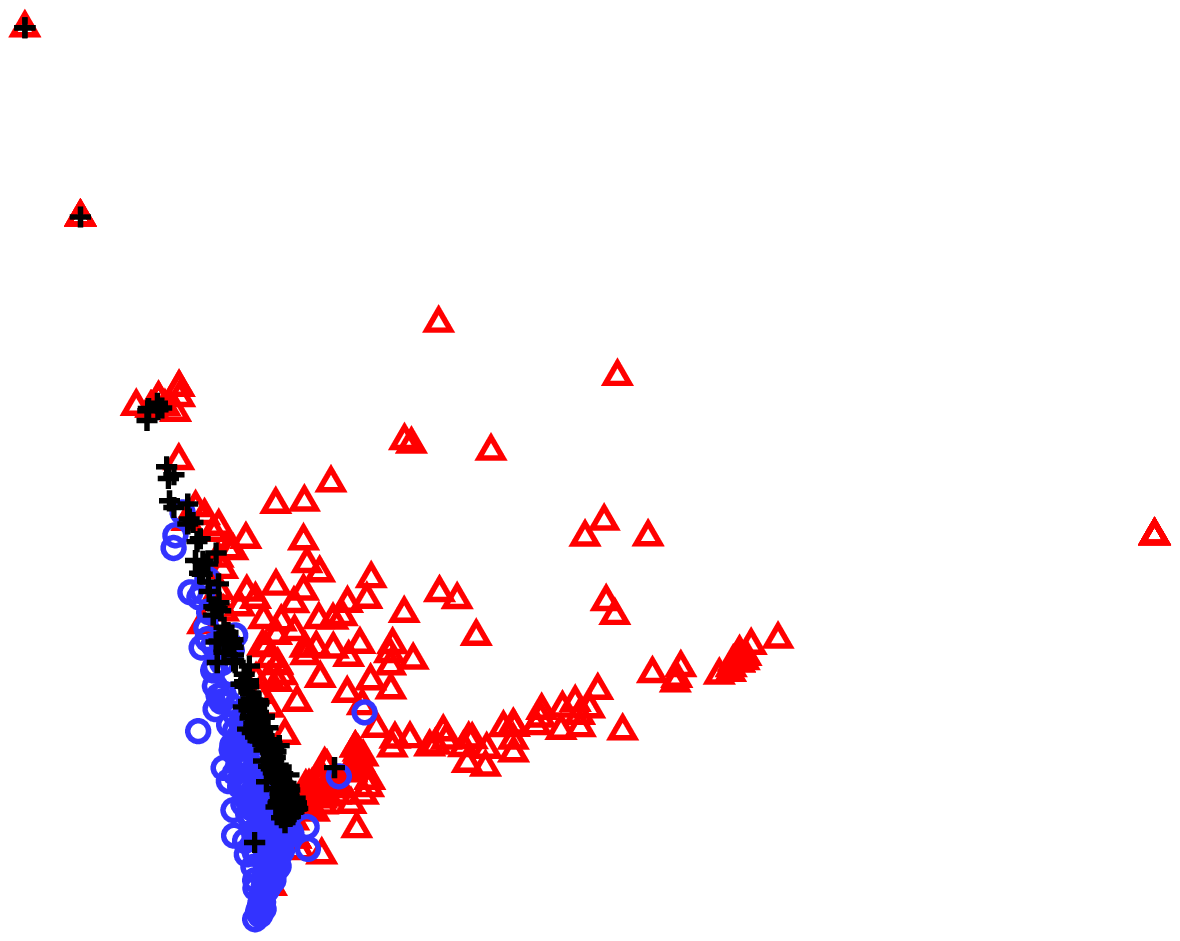}}} \hfil
\subfigure{\fbox{\includegraphics[scale=0.33]{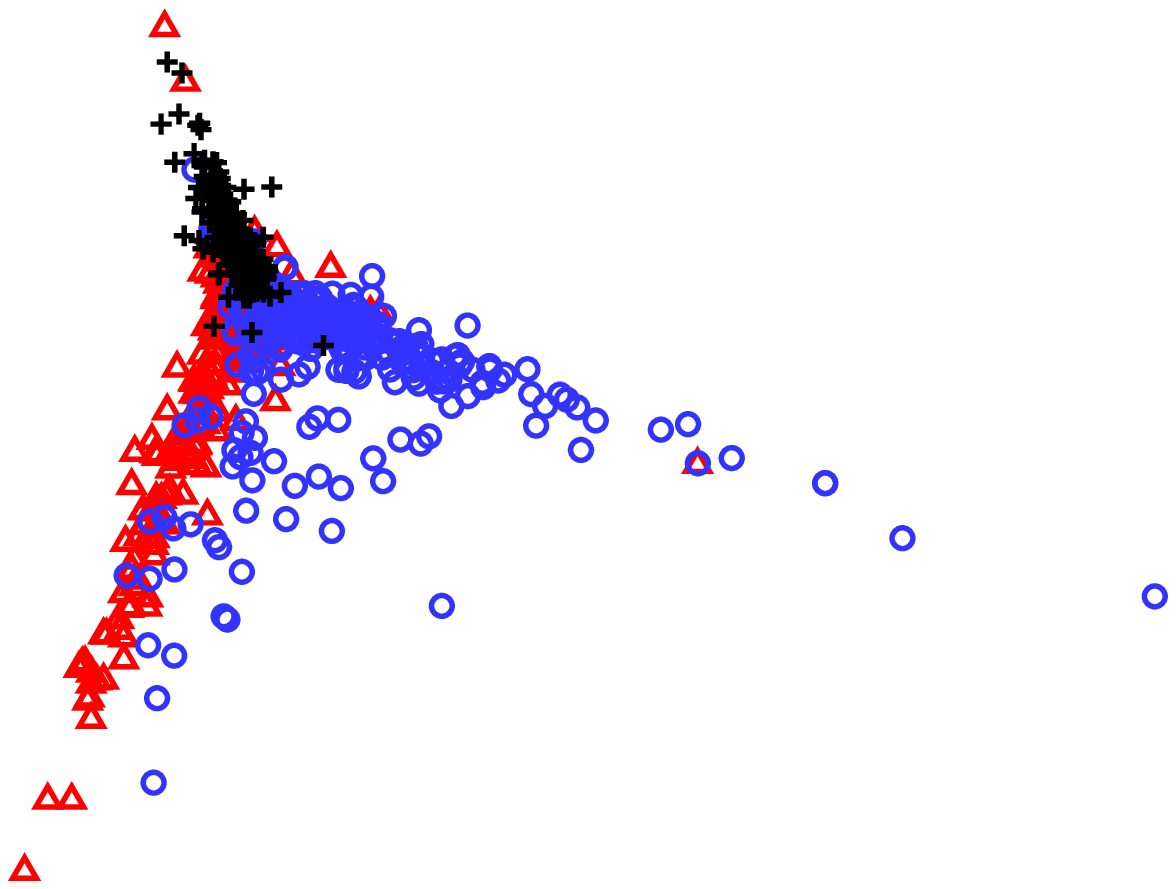}}} \hfil
\subfigure{\fbox{\includegraphics[scale=0.33]{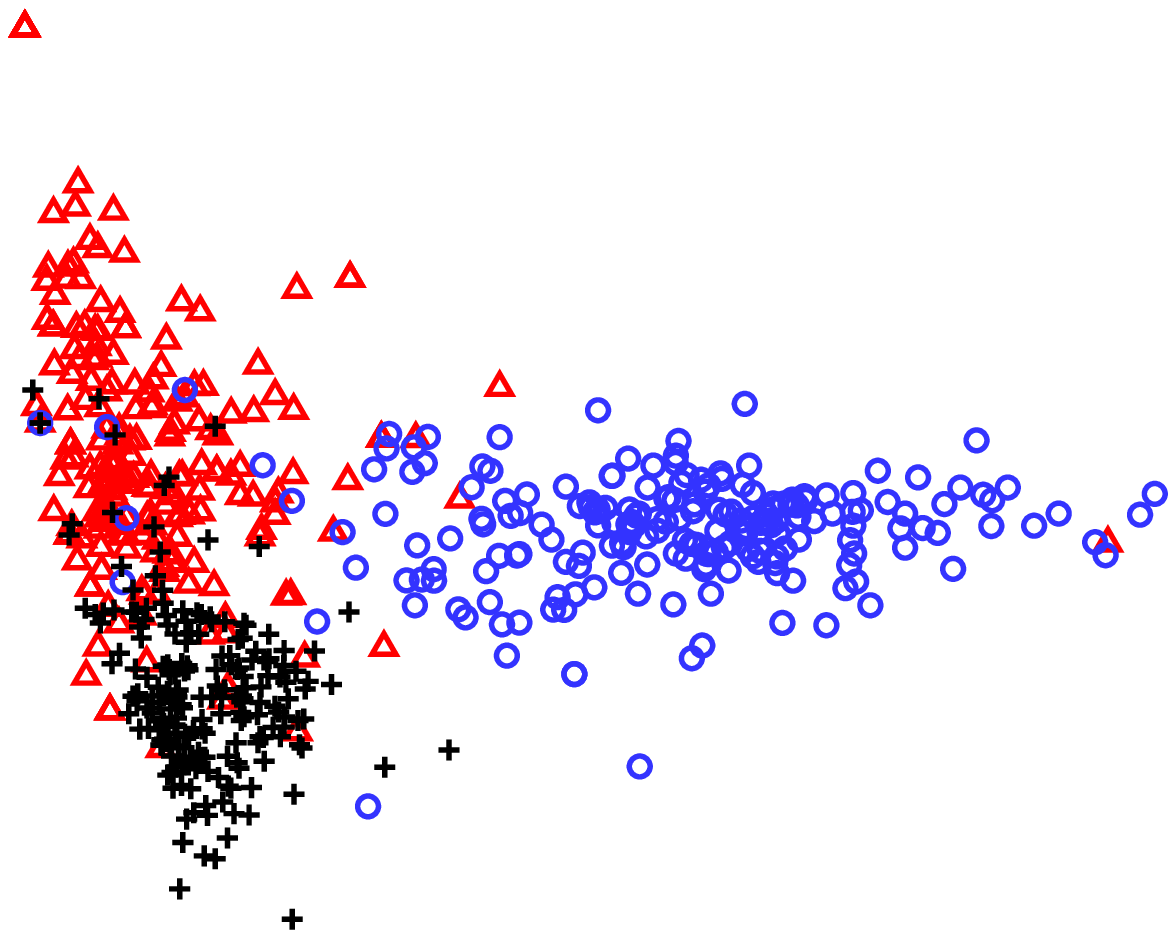}}}\\
\subfigure{\fbox{\includegraphics[scale=0.33]{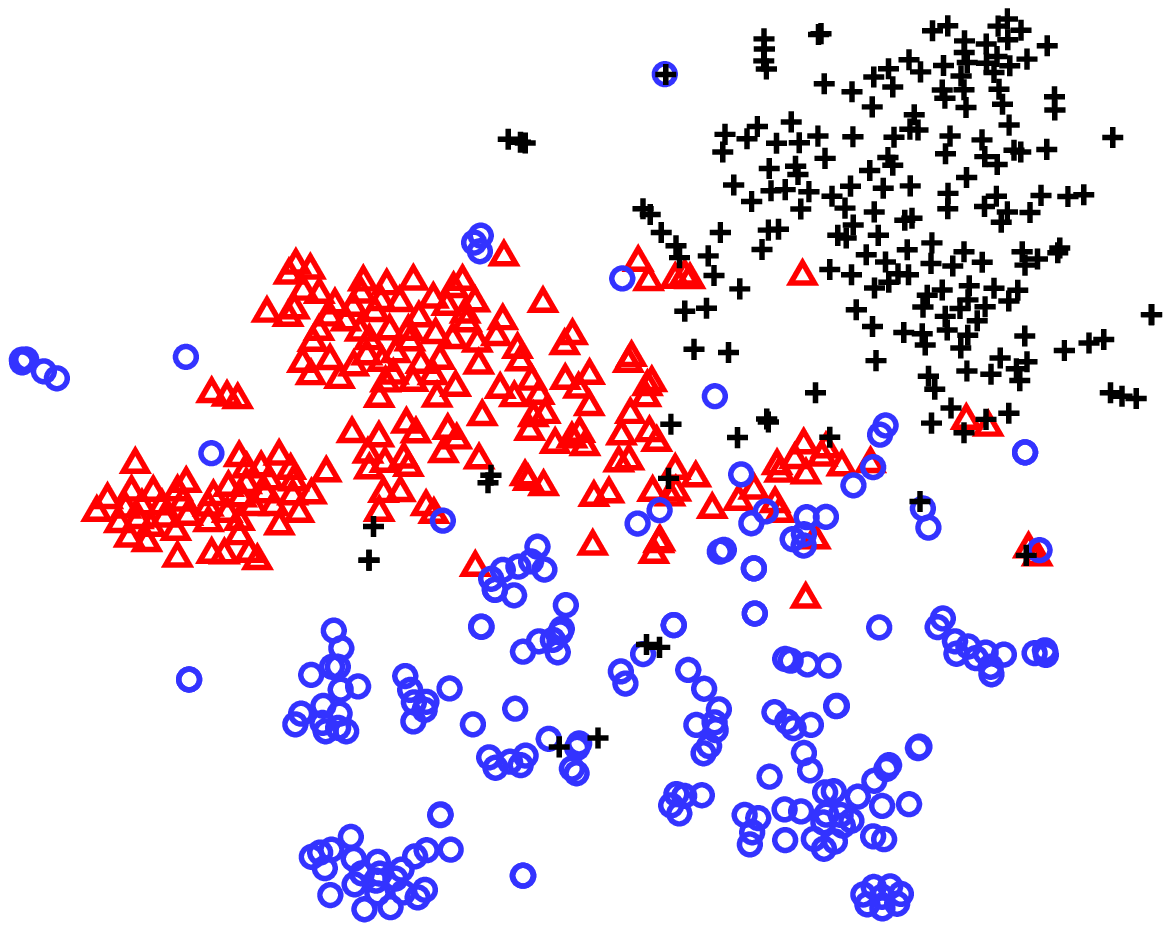}}} \hfil
\subfigure{\fbox{\includegraphics[scale=0.33]{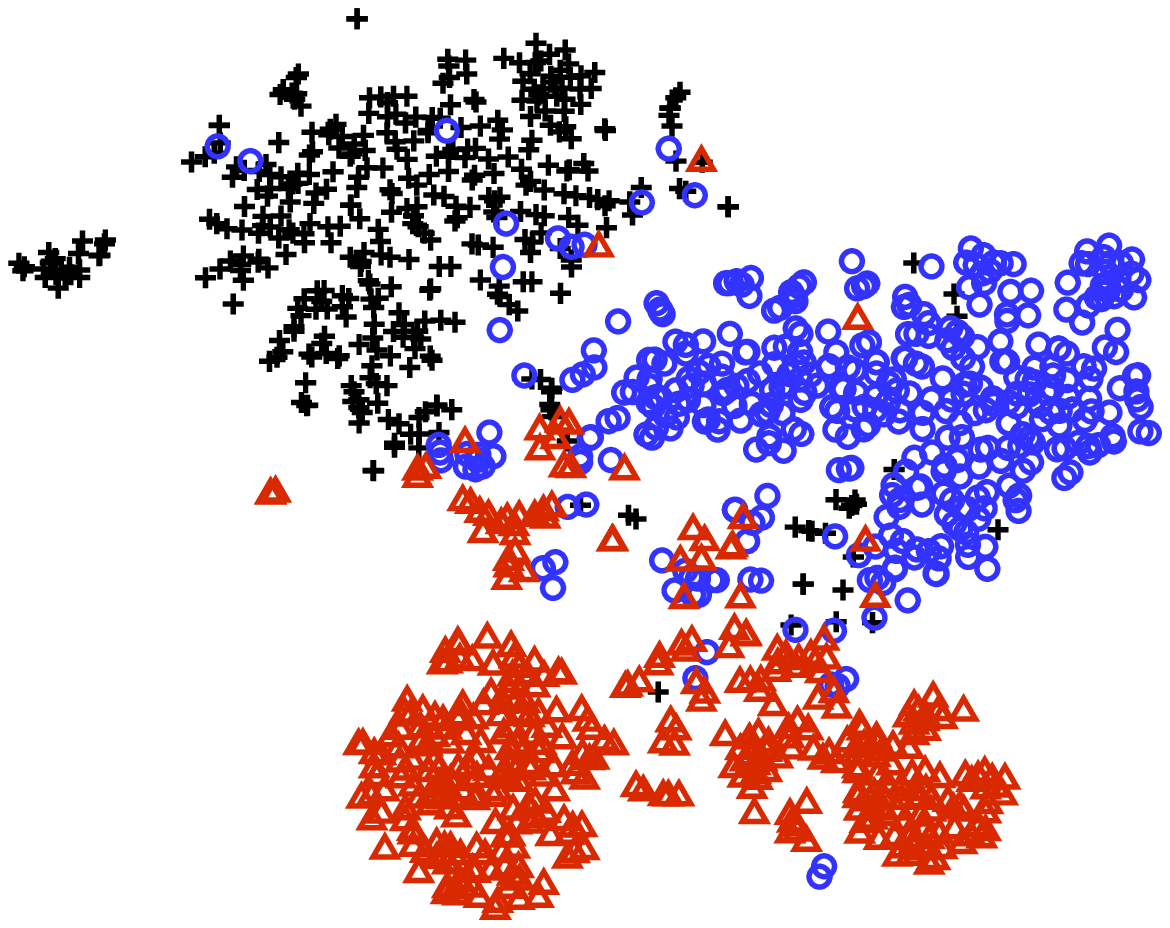}}} \hfil
\subfigure{\fbox{\includegraphics[scale=0.33]{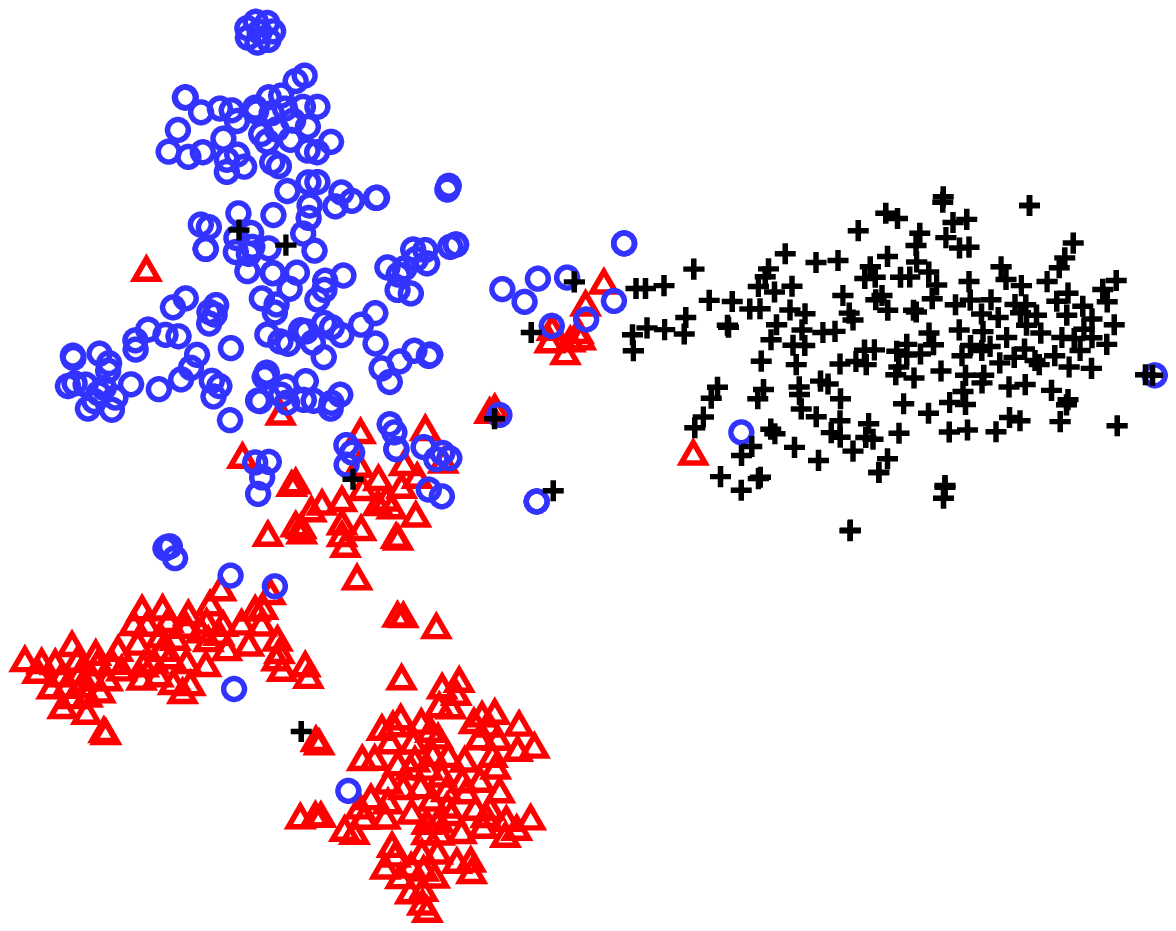}}}
\caption{ Qualitative evaluation of dimensionality reduction for the sentiment domain (top two rows) and the newsgroup domain (bottom two rows). The first and the third rows display PCA reduction while the second and the fourth display t-SNE. The left column correspond to no domain knowledge ($H=I$) reverting PCA and t-SNE to their original form. The middle column corresponds to manual specification (method A). The right column corresponds to contextual diffusion (method B). Different groups (sentiment labels or newsgroup labels) are marked with different colors and marks. \newline
In the sentiment case (top two rows) the graphs were rotated such that the direction returned by applying Fisher linear discriminant onto the projected 2D coordinates aligns with the positive x-axis. The bell curves are Gaussian distributions fitted from the x-coordinates of the     projected data points (after rotation). The numbers displayed in each sub-figure are computed from measure (iv).\newline
} \label{fig:newsgroupVis}
\end{figure*}

Figures~\ref{fig:newsgroupVis} displays qualitative and quantitative evaluation of PCA and t-SNE for the sentiment and newsgroup domains with standard $H=I$ geometry (left column), manual specification (middle column) and contextual diffusion (right column). Generally, we conclude that in both the newsgroup domain and the sentiment domain and both qualitatively and quantitatively (using the numbers in the top two rows), methods A and B perform better than using the original geometry $H=I$ with method B outperforming method A.

Tables~\ref{tab:sentiment}-\ref{tab:newsgroups} display two evaluation measures for different types of domain knowledge (see previous section). Table~\ref{tab:sentiment} corresponds to the sentiment domain where we conducted separate experiment for four movie critics. Table~\ref{tab:newsgroups} corresponds to the newsgroup domain where two tasks were considered. The first involving three newsgroups (classes comp.sys.mac.hardware, rec.sports.hockey and talk.politics.mideast) and the second involving four newsgroups (rec.autos, rec.motocycles, rec.sports.baseball and rec.sports.hockey). We conclude from these two figures that the contextual diffusion, Google $n$-gram, and Word-Net generally outperform the original $H=I$ matrix. The best method varies from task to task but the contextual diffusion and Google $n$-gram seem to have the strongest performance overall.

\begin{table}
  \centering \footnotesize
  \begin{tabular}{|c|cc|cc|cc|cc|}
    \hline
             & PCA (1)  & PCA (2)  & t-SNE (1) & t-SNE (2) \\\hline
    $H=I$    & 1.5391   & 1.4085   & 1.1649    & 1.1206\\
     B        & 1.2570   & \textbf{1.3036} & 1.2182 & 1.2331\\
    C        & \textbf{1.2023} & 1.3407 & \textbf{0.7844} & \textbf{1.0723}\\
    D        & 1.4475   & 1.3352   & 1.1762    & 1.1362\\
    \hline
\hline
        & PCA (1)          &  PCA (2)         &  t-SNE (1)       &  t-SNE (2)   \\ \hline
  $H=I$ & 0.8461           &  0.5630          &  0.9056          &  0.7281      \\
   B     & 0.7381           &  \textbf{0.6815} &  0.9110          &  0.6724      \\
  C     & 0.8420           &  0.5898          &  \textbf{0.9323} &  0.7359      \\
  D     & \textbf{0.8532}  &  0.5868          &  0.9013          &  \textbf{0.7728} \\ \hline
  \end{tabular}
\caption{Quantitative evaluation of dimensionality reduction for visualization for two tasks in the news article domain. The numbers in the top five rows correspond to measure (i) (lower is better), and the numbers in the bottom five rows correspond to measure (iii) ($k=5$) (higher is better). We conclude that contextual diffusion (B), Google $n$-gram (C), and Word-Net (D) tend to outperform the original $H=I$.} \label{tab:newsgroups}
\end{table}
\begin{table}
\centering
\footnotesize
\begin{tabular}{|c|c|c|c|}
    \hline
    $(\alpha_1,\alpha_2,\alpha_3,\alpha_4)$ & (i) & (ii)  & (iii) (k=5)  \\ \hline
    (1,0,0,0) & 0.5756 & -3.9334  & 0.7666  \\
    (0,1,0,0) & 0.5645 & -4.6966  & 0.7765 \\
    (0,0,1,0) & 0.5155 & -5.0154  & 0.8146 \\
    (0,0,0,1) & 0.6035 & -3.1154  & 0.8245 \\
    (0.3,0.4,0.1,0.2)  & \textbf{0.4735} & \textbf{-5.1154} &   \textbf{0.8976} \\ \hline
\end{tabular}\vspace{-0.05in}
\caption{ Three evaluation measures (i), (ii), and (iii) (see the beginning of the section for description) for convex combinations \eqref{eq:combination} using different values of $\alpha$. The first four rows represent methods A, B, C, and D. The bottom row represents a convex combination whose coefficients were obtained by searching for the minimizer of measure (iii). Interestingly the minimizer also performs well on measure (i) and more impressively on the labeled measure (iii).}
\label{tab:convcomb}
\end{table}

\begin{table*}
  \centering \footnotesize
  \begin{tabular}{|c|cc|cc|cc|cc|}
    \hline & \multicolumn{2}{|c|}{Dennis Schwartz} & \multicolumn{2}{|c|}{James Berardinelli} & \multicolumn{2}{|c|}{Scott Renshaw} & \multicolumn{2}{|c|}{Steve Rhodes}\\
          & PCA     & t-SNE    & PCA     & t-SNE   & PCA      & t-SNE   & PCA     & t-SNE\\\hline
    $H=I$ & 1.8625  & 1.8781   & 1.4704  & 1.5909  & 1.8047   & 1.9453  & 1.8013  & 1.8415\\
    A     & 1.8474  & 1.7909   & 1.3292  & 1.4406  & 1.6520   & 1.8166  & \textbf{1.4844}  & 1.6610\\
        B     & \textbf{1.4254} & \textbf{1.5809} & \textbf{1.3140} & \textbf{1.3276} & \textbf{1.5133} & \textbf{1.6097} & 1.5053 & \textbf{1.6145}\\
    C     & 1.6868  & 1.7766   & 1.3813  & 1.4371  & 1.7200   & 1.8605  & 1.7750  & 1.7979\\
    \hline \hline
    $H=I$ & 0.6404  & 0.7465   & 0.8481  & 0.8496  & 0.6559   & 0.6821  & 0.6680  & 0.7410 \\
    A     & 0.6011  & 0.7779   & \textbf{0.9224} & 0.8966 & 0.7424  & 0.7411 & \textbf{0.8350} & 0.8513 \\
        B     & \textbf{0.8831} & \textbf{0.8554} & 0.9188 & \textbf{0.9377} & \textbf{0.8215} & \textbf{0.8332} & 0.8124 & \textbf{0.8324} \\
    C     & 0.7238  & 0.7981   & 0.8871  & 0.9093  & 0.6897   & 0.7151  & 0.6724  & 0.7726 \\
    \hline
\end{tabular}
\caption{ Quantitative evaluation of dimensionality reduction for visualization in the sentiment domain. Each of the four columns corresponds to a different movie critic from the Cornell dataset (see text). The top five rows correspond to measure (i) (lower is better) and the bottom five rows correspond to measure (iii) ($k=5$, higher is better). Results were averaged over 40 cross validation iterations. We conclude that all methods outperform the original $H=I$ with the contextual diffusion and manual specification generally outperforming the others.} \label{tab:sentiment}
\end{table*}

\begin{figure*}[t]
\centering
\subfigure{\fbox{\includegraphics[scale=0.34]{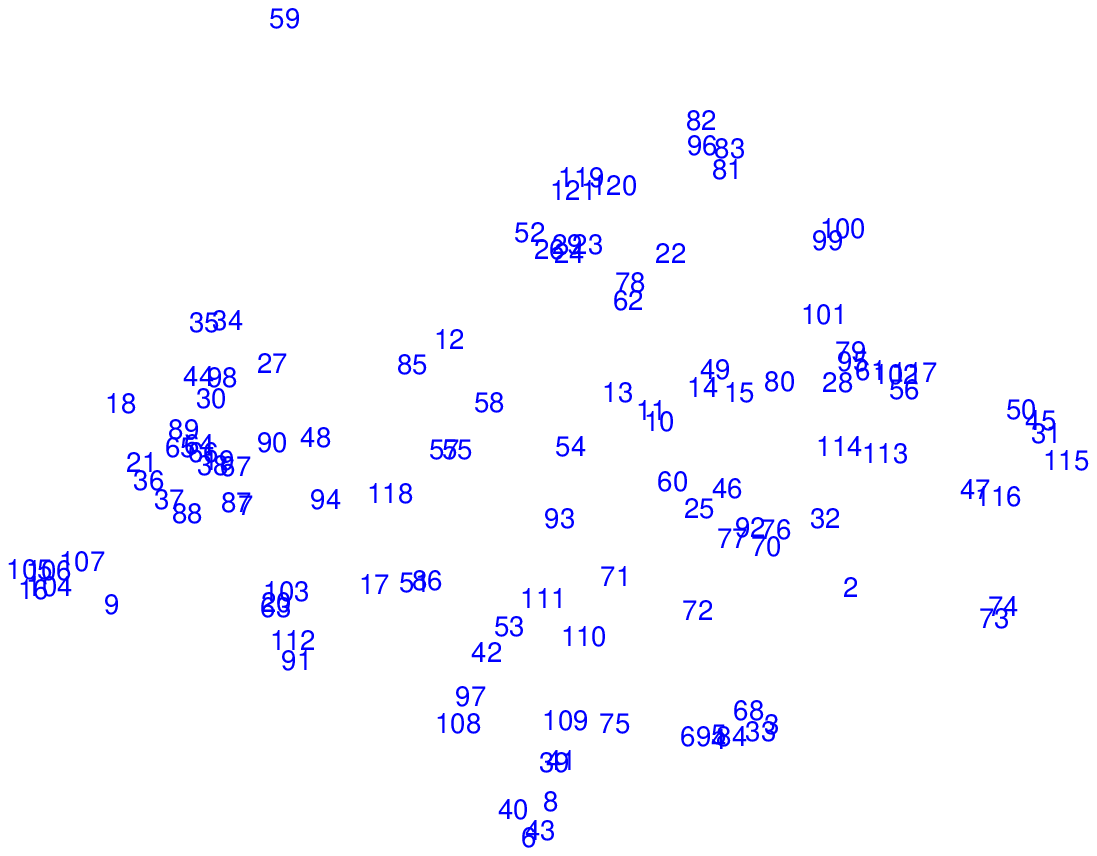}}} \hfil
\subfigure{\fbox{\includegraphics[scale=0.34]{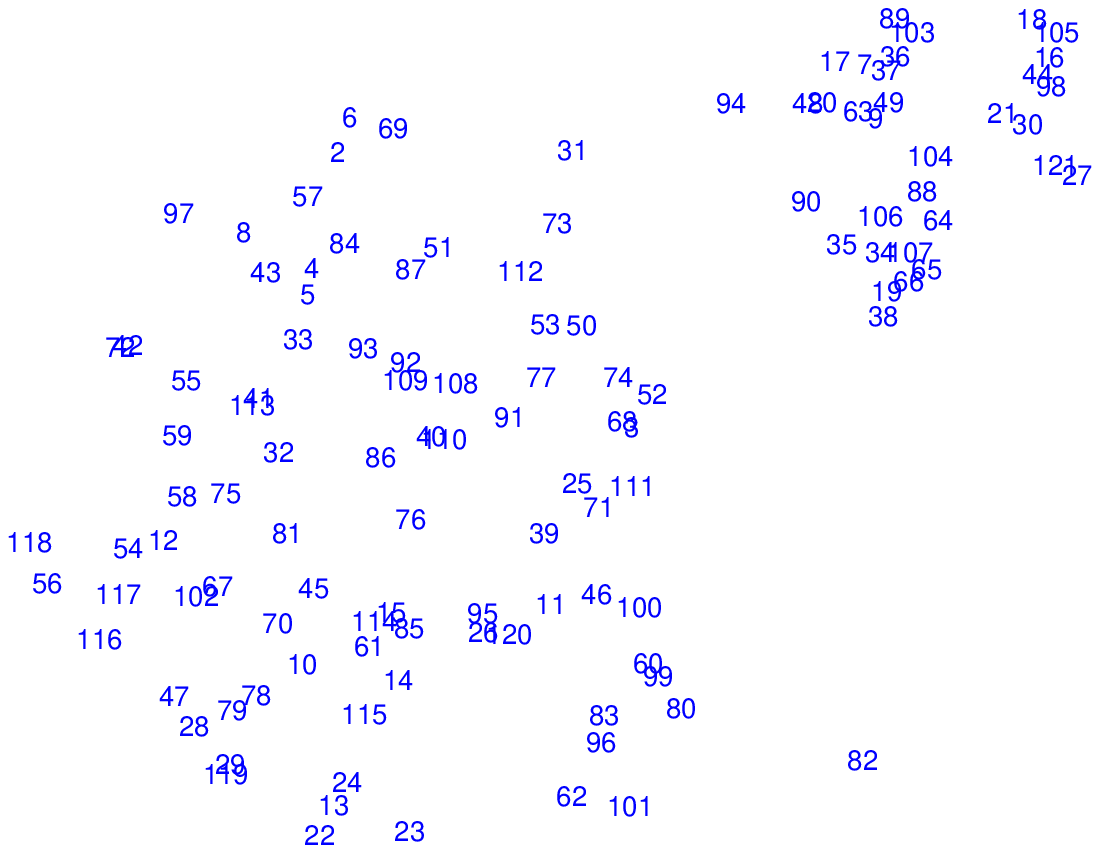}}} \hfil
\subfigure{\fbox{\includegraphics[scale=0.34]{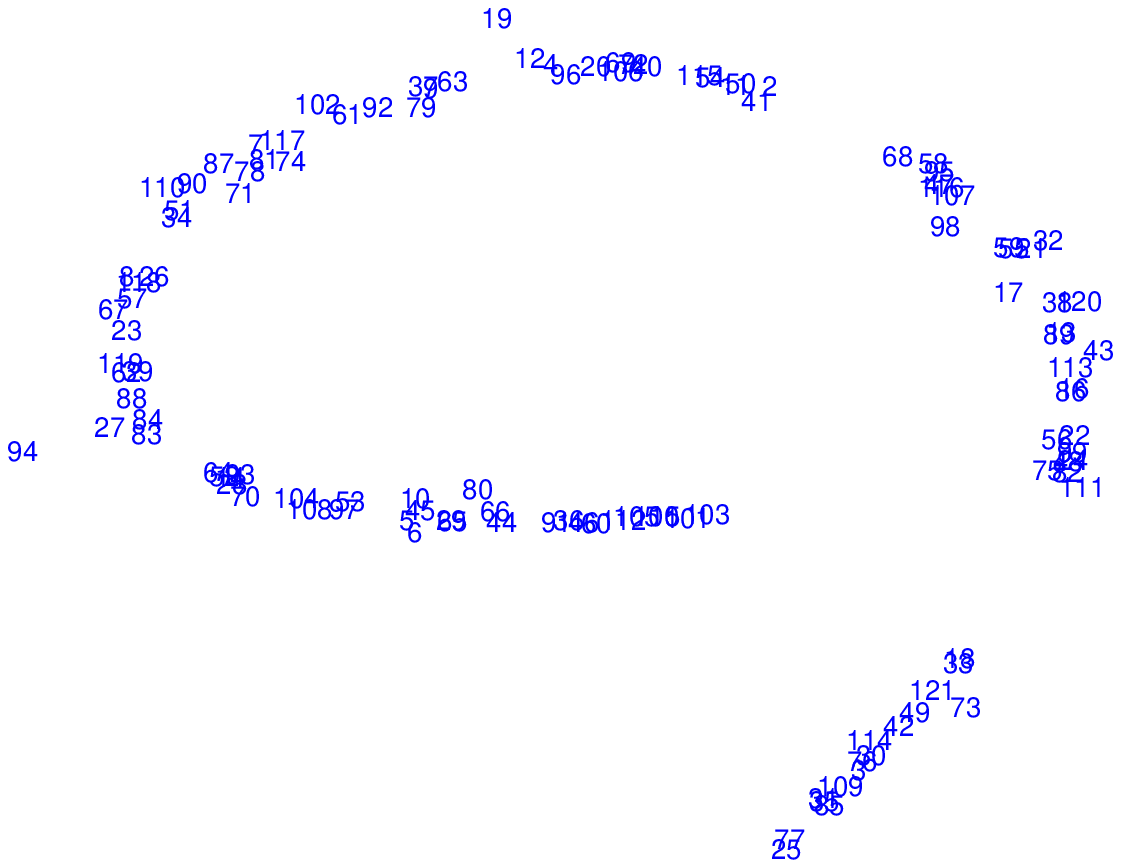}}}\\
\caption{Qualitative evaluation of dimensionality reduction for the ACL dataset using t-SNE. Left: no domain knowledge ($H = I$);  Middle: manual specification (method A); Right: contextual diffusion (method B). Each document is labeled by its assigned id from ACL anthology.} \label{fig:acl}
\end{figure*}

We also examined convex combinations
\begin{align} \label{eq:combination}
\alpha_1 H_A +\alpha_2 H_B +\alpha_3 H_C +\alpha_4 H_D
\end{align}
with $\sum\alpha_i=1$ and $\alpha_i\geq 0$. Table~\ref{tab:convcomb} displays three evaluation measures, the weighted intra-inter measure (i), the Davies-Bouldin index (ii), and the $k$-NN classifier ($k=5$) accuracy on the embedded documents (iii). The beginning of the section provides more information on these measures. The first four rows correspond to the ``pure'' methods A,B,C,D. The bottom row correspond to a convex combination found by minimizing the unsupervised evaluation measure (ii). Note that the convex combination found also outperforms A, B, C, and D on measure (i) and more impressively on measure (iii) which is a supervised measure that uses labeled data (the search for the optimal combination was done based on (ii) which does not require labeled data). We conclude that combining heterogeneous domain knowledge may improve the quality of dimensionality reduction for visualization, and that the search for an improved convex combination may be accomplished without the use of labeled data.

Finally, we demonstrate the effect of linguistic geometries on a new dataset that consists of all oral papers appearing in ACL 2001 -- 2009. For the purpose of manual specification, we obtain 1545 unique words from paper titles, and assign each word relatedness scores for each of the following clusters: morphology/phonology, syntax/parsing, semantics, discourse/dialogue, generation/summarization, machine translation, retrieval/categorization and machine learning. The score takes value from 0 to 2, where 2 represents the most relevant. The score information is then used to generate the transformation matrix $R$. We also assign each word an importance value ranging from 0 to 3 (larger the value, more important the word). This information is used to generate the diagonal matrix $D$. Figure \ref{fig:acl} shows the projection of all 2009 papers using t-SNE (papers from 2001 to 2008 are used to estimate contextual diffusion). The manual specification improves over no domain knowledge by separating documents into two clusters. By examining the document id, we find that all papers appearing in the smaller cluster correspond to either machine translation or multilingual tasks. Interestingly, the contextual diffusion results in a one-dimensional manifold.

\section{Discussion} \label{sec:discussion}
In this paper we introduce several ways of incorporating domain knowledge into dimensionality reduction for visualization of text documents. The novel methods of manual specification, contextual diffusion, Google $n$-grams, and Word-Net all outperform in general the original assumption $H=I$. We emphasize that the baseline $H=I$ is the one currently in use in most text visualization systems. The two reduction methods of PCA and t-SNE represent a popular classical technique and a recently proposed technique that outperforms other recent competitors (LLE, Isomap, MVU, CCA, Laplacian eigenmaps).

Our experiments demonstrate that different domain knowledge methods perform best in different situations. As a generalization, however, the contextual diffusion and Google $n$-gram methods had the strongest performance. We also demonstrate how combining different types of domain knowledge provides increased effectiveness and that such combinations may be found without the use of labeled data.

\bibliographystyle{plain} {\bibliography{../../common/groupPapers,../../common/externalPapers}}

\end{document}